\documentclass[sigconf,authorversion,nonacm]{acmart}

\AtBeginDocument{%
  \providecommand\BibTeX{{%
    \normalfont B\kern-0.5em{\scshape i\kern-0.25em b}\kern-0.8em\TeX}}}

\setcopyright{acmcopyright}
\copyrightyear{2021}
\acmYear{2021}
\acmDOI{10.1145/1122445.1122456}

\acmBooktitle{In Proceedings of the 15th International Conference on Web Search and Data Mining (WSDM'22), February 21-25, 2022, Arizona, USA.}

\usepackage{comment}
\usepackage{enumerate}
\usepackage[shortlabels]{enumitem}
\usepackage[singlelinecheck=false,justification=justified]{caption}
\usepackage{algorithm} 
\usepackage{algpseudocode} 
\usepackage{amsmath}

\usepackage{amssymb}
\usepackage{multirow}
\usepackage{float}
\usepackage{lipsum}

\usepackage{graphics}
\usepackage{tikz,pgfplots}
\usepackage{soul}
\usepackage{capt-of}
\usepackage{caption}
\usepackage{soul}
\usepackage{makecell}
\usepackage{bm}
\usepackage{dblfloatfix}
\usepackage{natbib}
\usepgfplotslibrary{groupplots}

\newcolumntype{Y}{>{\centering\arraybackslash}X}
\usepackage{tabularx}
\usepackage{amsthm}

\usepackage{underoverlap}

\usepackage{kotex}

\definecolor{customblue}{HTML}{20639B}
\definecolor{customred}{HTML}{ED553B}
\definecolor{customyellow}{HTML}{EB9605}
\definecolor{custompurple}{HTML}{9867C5}
\definecolor{customgreen}{HTML}{35B37E}

\definecolor{shcolor}{rgb}{0,0,1}

\begin{document}

\title{MOI-Mixer: Improving MLP-Mixer with Multi Order Interactions in Sequential Recommendation}

\author{Hojoon Lee}
\email{joonleesky@kaist.ac.kr}
\affiliation{%
  \institution{KAIST AI}
  \country{}
}

\author{Dongyoon Hwang}
\email{godnpeter@kaist.ac.kr}
\affiliation{%
  \institution{KAIST AI}
  \country{}
}

\author{Sunghwan Hong}
\email{sung_hwan@korea.ac.kr}
\affiliation{%
  \institution{Korea University}
  \country{}
} 

\author{Changyeon Kim}
\email{matthew.g@kakaocorp.com}
\affiliation{%
  \institution{KAKAO}
  \country{}
} 

\author{Seungryong Kim}
\email{seungryong_kim@korea.ac.kr}
\affiliation{%
  \institution{Korea University}
  \country{}
} 

\author{Jaegul Choo}
\email{jchoo@kaist.ac.kr}
\affiliation{
  \institution{KAIST AI}
  \country{}
} 

\renewcommand{\shortauthors}{Lee et al.}






\newcommand{\ourmodel}{MOI-Mixer}

\begin{abstract}
Successful sequential recommendation systems rely on accurately capturing the user’s short-term and long-term interest. Although Transformer-based models achieved state-of-the-art performance in the sequential recommendation task, they generally require \textit{quadratic} memory and time complexity to the sequence length, making it difficult to extract the long-term interest of users. 
On the other hand, Multi Layer Perceptrons (MLP)-based models,
renowned for their \textit{linear} memory and time complexity, have recently shown competitive results compared to Transformer in various tasks. 
Given the availability of a massive amount of the user's behavior history, the linear memory and time complexity of MLP-based models make them a promising alternative to explore in the sequential recommendation task.
To this end, we adopted MLP-based models in sequential recommendation but consistently observed that MLP-based methods obtain lower performance than those of Transformer despite their computational benefits.
From experiments, we observed that introducing explicit high-order interactions to MLP layers mitigates such performance gap.
In response, we propose the Multi-Order Interaction (MOI) layer, which is capable of expressing an arbitrary order of interactions within the inputs while maintaining the memory and time complexity of the MLP layer. 
By replacing the MLP layer with the MOI layer, our model was able to achieve comparable performance with Transformer-based models while retaining the MLP-based models' computational benefits. 
\end{abstract}

\maketitle
\begin{figure*}[h]
    \begin{center}
    \includegraphics[width=1.0\linewidth]{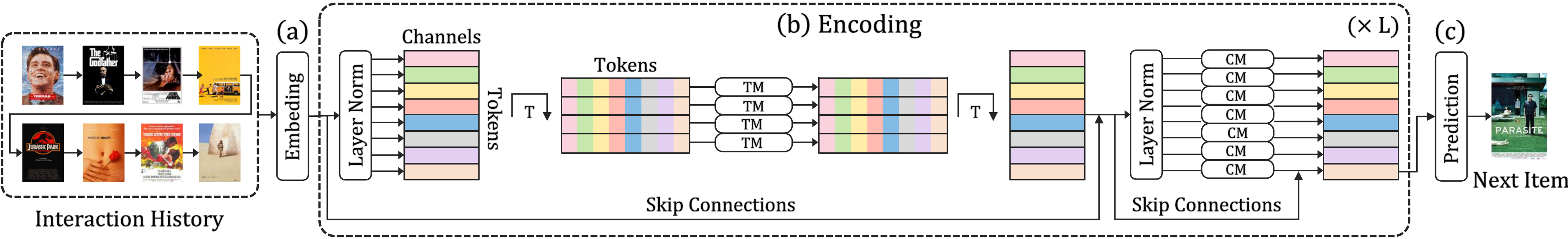}
    \end{center}
    \captionsetup{justification=centering}
    \caption{The overall architecture of \ourmodel{} which consists of 3 different modules: (a) an embedding; (b) an encoding; and (c) a prediction module. The goal is to predict the next interacted item based on the user's interaction history.}
    \label{fig:architecture}
\end{figure*}

\section{Introduction} \label{section:intro}
A sequential recommendation system is a core component for personalized recommendation services in Internet service platforms such as e-commerce, social network, and Over-the-top media.
Given a sequence of a user's historical behaviors, the prediction performance of a sequential recommendation system depends on accurately capturing the user's short- and long-term interest, referred as temporal dynamics.
To this end, early work utilized Markov Chains~\cite{fpmc} or Recurrent Neural Networks (RNN)~\cite{gru4rec} to represent the temporal dynamics within the user's historical behaviors.

Over the past decades, Internet service platforms accumulated a massive amount of user behavior history with extremely long lengths~\cite{ren2019lifelong, wu2021lima}. With the availability of such data, extracting users' long-term interest has become a key challenge for constructing an effective sequential recommendation system. 
However, RNN-based architectures struggle to propagate information through long sequences due to the vanishing gradient problem~\cite{RNN-Bengio}.
As an alternative, Transformer \cite{vaswani2017attention} was proposed to overcome the limitations of RNNs by using a self-attention layer. 
The self-attention layer constructs input-dependent attention weights which provide a global receptive field over the entire sequence and directly connect all the items (e.g., tokens) in the sequence. 
With such a global receptive field, Transformer effectively extracts the user's long-term interest and attains state-of-the-art performance in natural language processing~\cite{devlin2018bert} and sequential recommendation tasks~\cite{sasrec, bert4rec}.

Although the self-attention layer successfully extracts users' long-term interest, the attention weights make the memory and time complexity \textit{quadratic} to the sequence length and this restricts the applicability of Transformer-based models from processing extremely long sequence data. Recently, MLP-based architectures~\cite{touvron2021resmlp, liu2021pay_attention, tatsunami2021raftmlp, yu2021s2mlp}, including MLP-Mixer~\cite{tolstikhin2021mlpmixer}, have been proposed to replace the self-attention layer in Transformer with a simple MLP layer. 
The MLP layer integrates the items in the sequence with linear layers, successfully reducing the quadratic complexity to a linear complexity by substituting the self-attention layer.
Surprisingly, these architectures achieved competitive performance in comparison to Transformer for natural language processing. 
With the availability of the users' long behavior history, the linear computational cost of the MLP layers is indeed an attractive and promising direction to explore in the sequential recommendation systems.

From our extensive experiments, we observed that MLP-based methods performed relatively worse than Transformer-based methods for the sequential recommendation task despite its computational benefits. 
One of the convincing reasons could be the difference in the order of interactions a MLP layer and a self-attention layer exploit.
The MLP layer contains a first-order interaction term (e.g., $Wx$) whereas a self-attention layer contains a third-order interaction term (e.g., $qk^Tv$).
Although the first-order interaction term of the MLP layer is a universal function \cite{hornik1989multilayer}, several architectures that explicitly utilize complex interaction terms~\cite{hu2013cross, FiBiNet, ma2020memory} have shown their effectiveness for the sequential recommendation task.
Therefore, in this paper, we study the potentials of high-order interaction terms in the sequential recommendation task.

%

We propose a novel Multi-Order Interaction (MOI) layer designed to capture the desired $k$-th order interactions, while maintaining the linear computational cost of the MLP layer. 
By replacing MLP layers in MLP-Mixer with our MOI layers, we named our model as \ourmodel{}.
Note that MLP-Mixer is a special case of \ourmodel{} since the MOI layer with interaction order at $k=1$ is identical to the MLP layer.
\ourmodel{} alternates between (i) a token-mixing layer, which integrates the representations across tokens, and (ii) a channel-mixing layer, which integrates the representations within each token. 
We conducted extensive experiments on various datasets and observed that high-order interactions for the channel-mixing operation improves the performance of \ourmodel{}.

In summary, our contributions are threefold as follows:
\begin{itemize}
    \item To the best of our knowledge, we are the first to apply various state-of-the-art MLP-based architectures in sequential recommendation systems.
    \item We propose a novel Multi-Order Interaction (MOI) layer, which effectively expresses the arbitrary order of interaction within the inputs and stacked them to construct MOI-Mixer. 
    \item Through extensive experiments, we observe that MOI-Mixer surpasses existing MLP-based architectures, which validates the effectiveness of the high-order interactions in the sequential recommendation tasks. 
\end{itemize}

\section{Related Work} \label{section:relatedwork}

\subsection{Sequential Recommendation}
Sequential recommendation systems are based on the user's previous sequential behaviors to anticipate their future behavior. 
To address this problem, early work such as FPMC~\cite{fpmc} attempts to learn user-specific transition matrices using Markov chains to model sequential patterns.  
Inspired by neural models' strong representation capabilities, deep learning-based approaches~\cite{caser, gru4rec, Wu2018RNN, hidasi2016sessionbased} including recurrent and convolution neural networks have been proposed to learn complicated user behaviors and effectively capture long-term dependencies. 
Recently, Transformer~\cite{vaswani2017attention} directly connects all the items in the sequences and successfully extracts the user's long-term preferences. These methods achieved state-of-the-art performance in sequential recommendation~\cite{sasrec, bert4rec}.

\subsection{MLP-based Architectures}
Transformer architecture with the self-attention layer has attained state-of-the-art performance in various tasks including computer vision~\cite{dosovitskiy2020vit, cho2021semantic} and natural language processing~\cite{devlin2018bert}.
However, recent studies have shown that MLP-based architectures~\cite{tolstikhin2021mlpmixer, touvron2021resmlp, liu2021pay_attention, tatsunami2021raftmlp, yu2021s2mlp} can achieve competitive performance against Transformer for both computer vision and natural language processing. 
These MLP-based models have similar macro-level architecture to Transformer but differ in the micro-component designs. MLP-Mixer~\cite{tolstikhin2021mlpmixer} and ResMLP~\cite{touvron2021resmlp} simply replaced the self-attention layer with the MLP layer and obtained competitive performance on image classification benchmarks. gMLP~\cite{liu2021pay_attention} used the gated version of the MLP layer, further enhancing the performance of the MLP-based models. 


\section{Preliminary} \label{section:rq1}

This section describes the problem settings of sequential recommendation and the macro-level architecture of \ourmodel{}.

\subsection{Problem Formulation}
Sequential recommendation aims to model the users' item preference by predicting the next item given the past interacted items (i.e., watched movies)~\cite{sasrec, bert4rec}. 
Formally, given a sequence of users' interacted items with length $s$, at time step $t$ (i.e., interacted items from time step $t-s+1$ to $t$), the objective is to predict the next item at time step $t+1$ that the user is likely to interact with.

\subsection{Overall Architecture}

Fig.~\ref{fig:architecture} depicts the overall architecture of \ourmodel{}. \ourmodel{} closely follows the macro-structure of MLP-Mixer and ResMLP~\cite{tolstikhin2021mlpmixer, touvron2021resmlp}, which is composed of three different modules: (a) an embedding, (b) an encoding, and (c) a prediction module. 

The embedding module in Fig.~\ref{fig:architecture}(a) takes a sequence of $s$ number of items in the user's item interaction history as an input, and each item is projected onto a $c$-dimensional token resulting in an input matrix of $X \in \mathbb{R}^{s \times c}$. 
Note that while previous architectures with self-attention layers utilized positional embedding to provide either absolute or relative positional information in a given sequence to cope with the permutation invariance property~\cite{sasrec,bert4rec}, \ourmodel{} does not require the positional embedding since the encoding layer within our proposed architecture is already sensitive to the order of the tokens, allowing inherent learning of position of each sequence.

Then, in Fig.~\ref{fig:architecture}(b), the embedded input $X$ is fed into the encoder module which consists of a stack of identical blocks where each block is composed of two mixing operations.
The first operation conducts token-wise mixing, which we call a token-mixing layer, $TM(\cdot)$. It acts identically on the columns of $X$ to capture the interaction between the tokens within a channel which maps $\mathbb{R}^{s} \mapsto \mathbb{R}^{s}$. 
Then, the results are fed to a channel-mixing sub-layer, $CM(\cdot)$, which acts on the rows of $X$ to capture the interaction between the channels within a token, which maps $\mathbb{R}^{c} \mapsto \mathbb{R}^{c}$.
Following~\cite{bert4rec, tolstikhin2021mlpmixer, touvron2021resmlp}, standard architectural components such as residual connections~\cite{resnet} and layer normalization~\cite{layernorm} are utilized to stabilize the training process. 
Omitting the layer indices for brevity, the encoder layer of \ourmodel{} is written as
\begin{equation}
\begin{split}
    Y_{*,i} &= X_{*,i} + TM(\text{LayerNorm}(X)_{*,i}), \quad  \text{for} \; i = 1...c, \\ 
    Z_{j,*} &= Y_{j,*} + CM(\text{LayerNorm}(Y)_{j,*}), \quad  \text{for} \; j = 1...s,
\end{split}
\end{equation}
where $Y, Z \in \mathbb{R}^{s \times c}$ indicates the output of the token-mixing layer and the channel-mixing layer, respectively.

Once the input undergoes the encoding module, the encoded representation of the last token $X_s \in R^c$ in the sequence is passed to the prediction module for the final classification problem, as shown in Fig.~\ref{fig:architecture}(c). The prediction module is composed of a two-layer feed-forward network with GELU activation~\cite{hendrycks2020gelu}, followed by the softmax function as done in~\cite{bert4rec}. As a result, we obtain a probability distribution over the items for the next token.

The difference between the MLP-Mixer~\cite{tolstikhin2021mlpmixer} and ours lies on the operations of the token-mixer $TM(\cdot)$ and the channel-mixer $CM(\cdot)$. 
The MLP-Mixer utilized a MLP layer for $TM(\cdot)$ and $CM(\cdot)$, capturing the first-order interactions between the tokens and the channels. However, \ourmodel{} applies the MOI layer, capturing the multi-order interactions between the tokens and the channels.
\section{Proposed Method} \label{section:rq1}

Here, we explain our MOI (Multi-Order Interaction) layer designed to capture the multi-order interaction over the input features.


\subsection{Parameterized Multi-Order Interaction}
Our goal is to learn a weight tensor $\mathcal{T}$ which linearly combines the $k$-th order of interaction between $k$ different inputs. Let us denote the input vectors as $x_1 \in \mathbb{R}^{d_1}, \cdots, x_k\in \mathbb{R}^{d_k}$, the number of the hidden dimensions as $d_1, \cdots, d_k$, the desired order of interaction as $k$ and the output dimension as $h$.
In addition, we define the learnable tensor $\mathcal{T} \in \mathbb{R}^{{\scriptstyle d_1 \times \cdots \times d_k} \times h} $, which outputs $h$ different $k$-th order interactions over the input features denoted by $z \in \mathbb{R}^{h}$. Specifically, $z$ is defined as
\begin{equation}
    z^T = (((\mathcal{T} \times_1 x_1) \times_2 x_2) \times_3 \cdots \times_k x_k), 
\label{eq:moi}
\end{equation}
where operator $\times_i$ denotes the $i$-th mode tensor product.
\begin{figure}[h]
    \begin{center}
    \includegraphics[width=1.0\linewidth]{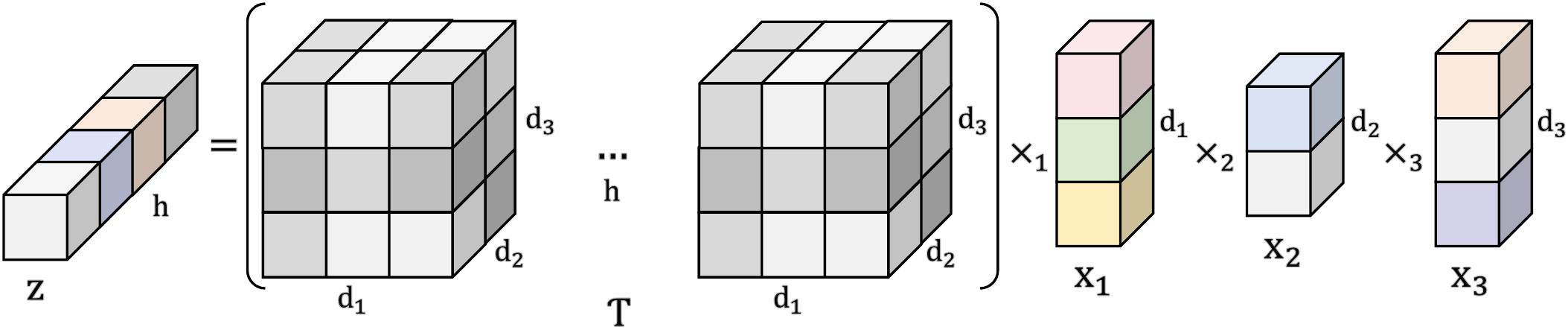}
    \end{center}
    \captionsetup{justification=centering}
    \caption{An illustrative example of the Parameterized Multi-Order Interaction for the interaction order $k=3$.}
    \label{fig:attention_illustration_pdf}
\end{figure}

\subsection{Low-Rank Approximation}
Although the tensor $\mathcal{T}$ could capture all possible $k$-th order interactions, learning such a large tensor $\mathcal{T}$ gets practically infeasible due to the exponential number of parameters as mode $k$ increases. 
To reduce the computational cost, we employ PARAFAC decomposition~\cite{kiers2000parafac} to perform a low-rank approximation of the tensor. Note that we denote $R$ as the rank of the decomposed matrices which controls a trade-off relationship between the decomposition rate and the computational cost.
Here, we set as $R=1$ and conduct the PARAFAC decomposition for the tensor $\mathcal{T}$, i.e.,
\begin{equation}
    \mathcal{T} \approx (((\mathcal{G} \times_1 W_1) \times_2 W_2) \times_3 \cdots \times_k W_k), 
\label{eq:parafac}
\end{equation}
where $\mathcal{G} \in \mathbb{R}^{\underbracket{\scriptstyle h \times h \times \cdots \times h}_{k+1}}$ is a rank-1 weighting tensor and $W_1 \in \mathbb{R}^{d_1 \times h}, \cdots, W_k \in \mathbb{R}^{d_k \times h}$ are the learnable factor matrices~\cite{do2019compact}.

Then, leveraging Eq.~\eqref{eq:parafac}, we can re-write the output $z$ as
\begin{equation}
    z^T \approx (((\mathcal{G} \times_1 x_1W_1) \times_2 x_2W_2) \times_3 \cdots \times_k x_kW_k). 
\label{eq:moi_parafac}
\end{equation}

Following~\cite{kolda2009tensor}, the result obtained from Eq.~\eqref{eq:moi_parafac} can be approximated by the Hadamard products without the presence of the rank-1 tensor $\mathcal{G}$. Hence the output $z$ can be approximated as
\begin{equation}
    z \approx (W_1^Tx_1 \odot \cdots \odot W_k^Tx_k), 
\label{eq:moi_hadamard}
\end{equation}
where $\odot$ indicates the Hadamard product.

In Eq.~\eqref{eq:moi_hadamard}, the weight matrices $W_1,...,W_k$ can have their own bias vectors $b_1, ..., b_k \in \mathbb{R}^{h}$. 
With the bias vectors, the output $z$ can represent all terms of the interactions less than or equal to $k$, i.e.,
\begin{equation}
\begin{split}
    z \approx & (W_1^Tx_1 + b_1) \odot \cdots \odot (W_k^Tx_k + b_k) \\
    = & (\mathbf{Wx} + \textstyle\sum_{i=1}^{i=k} \mathbf{Wx} \oslash W_i^Tx_i \cdot \text{diag}(b_i)\\
    &+ \textstyle\sum_{i=1}^{i=k}\sum_{j=i+1}^{j=k}\mathbf{Wx} \oslash (W_i^Tx_i \odot W_j^Tx_j) \cdot \text{diag}(b_i \odot b_j)\\
    &+ \cdots + (b_1 \odot \cdots \odot b_k)),
\end{split}
\end{equation}
where $\mathbf{Wx} = (W_1^Tx_1 \odot \cdots \odot W_k^Tx_k) $ and $\oslash$ indicates the Hadamard division.

\subsection{Multi-Order Interaction Layer}

The interaction model in Eq.~\eqref{eq:moi_hadamard} outputs $h$ different features of the $k$-th order interactions.  
We then introduce an additional linear layer $W_o \in \mathbb{R}^{h \times d}$ after the Hadamard product to not only aggregate, but also map the output $z$ onto the identical dimension as input $x$ for residual connections.

To further enhance the representative capacity of the model, we apply a non-linear activation function immediately after the input vectors, as in~\cite{kim2016hadamard}. 
However, the Hadamard product over the multiple inputs may induce the explosion of the output values. Therefore, to suppress the large output values, we exploit an additional layer normalization after the product, inspired by~\cite{Zheng2019finegrainedbilinear, pmlr2020lowrankbilinear}.

\begin{figure}[h]
    \begin{center}
    \includegraphics[width=1.0\linewidth]{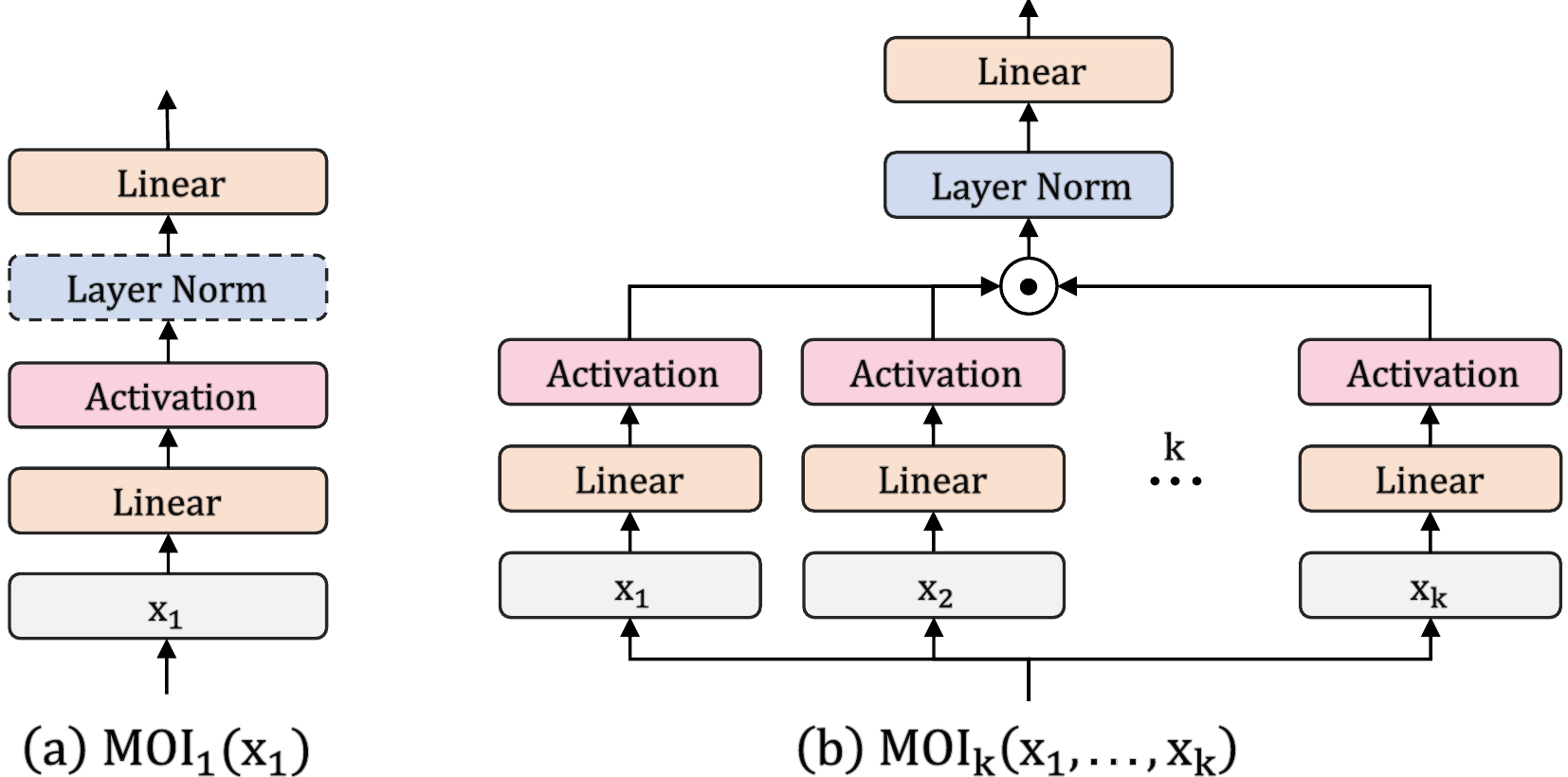}
    \end{center}
    \captionsetup{justification=centering}
    \caption{Our proposed MOI layer that captures the multi-order interaction between the input features.}
    \label{fig:model}
\end{figure}

To sum up, as illustrated in Fig.~\ref{fig:model}, we define our MOI (Multi-Order Interaction) Layer at the interaction order $k$ as
\begin{equation}
    MOI_k(x_1,..., x_k) = W_o^T(\text{LayerNorm}(\sigma(W_1^Tx_1) \odot \cdots \odot \sigma(W_k^Tx_k))),
\end{equation}
where $\sigma$ is an element-wise non-linear function (i.e., GELU~\cite{hendrycks2020gelu}).


\subsection{Special cases of MOI layer}
We now discuss the relationship between the MOI layer and existing layers with respect to the order $k$. Throughout the section, we omit the normalization layer for brevity.

\noindent
\textbf{MLP Layer} For the case where the MOI layer's order $k$ is set to $1$, it is identical to the MLP layer as shown in Fig.~\ref{fig:model}(a), i.e., 
\begin{equation}
    MOI_1(x) = W_o^T(\sigma(W_1^Tx)),
\end{equation}
where $x\in \mathbb{R}^{c}$ denotes the input vector, $W_1 \in \mathbb{R}^{c \times h}$ and $W_o \in \mathbb{R}^{h \times c}$ denote the fully connected layers with a hidden dimension $h$, and $\sigma$ is the activation function.
Note that we do not need the normalization layer due to the absence of the Hadamard product.

\noindent
\\
\textbf{Bilinear Pooling Layer} For the case where the order of MOI layer is $k=2$, the MOI layer is identical to the low-rank bilinear pooling layer proposed in~\cite{kim2016hadamard, kim2018bilinearattention} which is defined as
\begin{equation}
    MOI_1(x, y) =  W_o^T(\sigma(W_1^Tx) \odot \sigma(W_2^Ty)),
    \label{eq:bilinear}
\end{equation}
where $x, y \in \mathbb{R}^c$ are input vectors, $W_1, W_2 \in \mathbb{R}^{c \times h}$ are weight matrices, $\sigma$ is the activation and $W_o\in \mathbb{R}^{h \times c}$ is a pooling matrix.

\subsection{MOI-Mixer}

The final architecture of our proposed MOI-Mixer$_{(k_s, k_c)}$ is composed of two MOI layers, one for the token-mixing $TM(\cdot)$ layer, where $k_s$ indicates the order of token interaction and one for the channel-mixing $CM(\cdot)$ layer, where $k_c$ indicates the order of channel interaction. Since our goal is to introduce an explicit high-order term of the given input, we set all inputs $x_1, \cdots, x_k$ as a single input $x$ in the MOI layer.
We further denote the hidden dimension of $TM(\cdot)$ and $CM(\cdot)$, respectively, with $d_s$ and $d_c$.

\subsection{Relation to Previous Models}
In this section, we give a detailed  comparison between \ourmodel{} and relevant prior models~\cite{tolstikhin2021mlpmixer, touvron2021resmlp, liu2021pay_attention}.

\noindent
\\
\textbf{Relation to MLP-Mixer, ResMLP} : The overall structure of \ourmodel{} can easily be modified to MLP-Mixer~\cite{tolstikhin2021mlpmixer} and ResMLP~\cite{touvron2021resmlp}, the simplest MLP-based architecture, by simply setting the order of the MOI-Mixer$_{(k_s, k_c)}$ as $k_s=1$ and $k_c=1$ for the token-mixing $TM(\cdot)$ and the channel-mixing $CM(\cdot)$ layer. Note that ResMLP~\cite{touvron2021resmlp} requires an additional affine layer before and after the token-mixing layer for normalization.

\noindent
\\
\textbf{Relation to gMLP} : Different from the MLP-Mixer architecture, gMLP combines the token-mixing and the channel-mixing in a single module which can be written as:
\begin{equation}
    Z = (\sigma(XW_{1}) \odot W_s\sigma(XW_{2}))W_o,
    \label{eq:bilinear}
\end{equation}
where $X \in \mathbb{R}^{s \times c}$ is the input matrix, $W_1, W_2 \in \mathbb{R}^{c \times h}$ are channel-mixing weights, $W_s \in \mathbb{R}^{s \times s}$ is a token-mixing weight, and $W_o \in \mathbb{R}^{h \times c}$ is the output matrix.

By removing the token-mixing weight $W_s$, gMLP is identical to the channel-mixing layer $CM(\cdot)$ of MOI-Mixer with the order $k_c=2$. 
Since the token's interaction is modeled by a simple linear layer $W_s$, gMLP resembles \ourmodel{} when the interaction order of the token-mixer is $k_s=1$ and channel-mixer is $k_c=2$.
However, \ourmodel{} differs from the gMLP where the gMLP's computational costs of $W_s$ is quadratic to the sequence length which may not be a practical choice for the long sequence dataset.

\section{Experiments} \label{section:rq2}
In this section, we first provide our experimental settings and report the results with detailed analysis on standard benchmark datasets. Specifically, Section~\ref{section:task_settings} briefly introduces the datasets and evaluation protocols. Section~\ref{section:analysis_interaction_order} analyzes the effects of the interaction order. Then, Section~\ref{section:small_scale} compares \ourmodel{} and its variants with state-of-the-art recommendation architectures to verify the effectiveness of high-order interactions. Lastly, Section~\ref{section:ablation} conducts an ablation study for an in-depth analysis of of the proposed method.

\subsection{Experimental Setup} \label{section:task_settings}

\noindent
\textbf{Datasets} We evaluate our model on five sequential recommendation datasets where the sparsity and domain varies significantly. 
\begin{itemize}
    \setlength\itemsep{0.5em}
    \item \textit{Beauty}: This is a dataset introduced in~\cite{mcauley2015image}, containing series of product reviews crawled from the \textit{Amazon.com}. This dataset is known for its high sparsity. 
    \item \textit{Steam}: This is a dataset collected from the online game distribution platform, \textit{Steam}, which was first introduced in~\cite{sasrec}.
    \item \textit{MovieLens}: This is a widely-used benchmark dataset in recommendation systems containing each user's ratings for movies. We used MovieLens-1m and MovieLens-20m, where 1m and 20m indicate the number of interactions, respectively.
    \item \textit{XLong}: This is a dataset introduced by~\cite{ren2019lifelong}, which is the collection of the click logs of Alibaba's e-commerce platform, containing particularly long sequences of data.
\end{itemize}

\begin{table}[h]
\caption{Statistics of the processed dataset. The avg. int. denotes the average number of interacted items per user.}
\begin{tabular}{l l c c c c c}
\toprule
&\\[-3ex]               
Type        & Dataset   & Users  & Items   & Total int. (M)    & Avg. int.  \\[0.2ex]
\hline
&\\[-2ex]
\multirow{3}{*}{Small}
& Beauty            & 40,226       & 54,542      & 0.4              & 8.8            \\
& Steam             & 281,428      & 13,044      & 3.5              & 12.4           \\
& ML-1m             & 6040         & 3,416       & 1.0              & 163.5          \\
Large & ML-20m      & 138,493      & 26,774      & 20.0             & 144.4          \\
Long  & XLong       & 20,000       & 747,460     & 15.9             & 794.2          \\[0.2ex]
\bottomrule
\end{tabular}
\label{table:dataset}
\end{table}

We followed the data pre-processing procedure from~\cite{sasrec, bert4rec}, where all reviews and ratings are regarded as implicit feedback. Afterward, interacted items are grouped by users and ordered by timestamps. We only kept users and items that have at least five interactions to maintain the quality of the dataset. Detailed statistics of the processed datasets are outlined in Table~\ref{table:dataset}.

\noindent
\\
\textbf{Baselines} To verify the effectiveness of our model, we compare our method with well-known sequential recommendation baselines including POP, NCF~\cite{ncf}, GRU4Rec$^{+}$~\cite{gru4rec}, HPMN~\cite{ren2019lifelong} and BERT4Rec~\cite{bert4rec}. POP is one of the simplest baselines which recommends items based on rank-based popularity. NCF models the interactions between users and items via matrix factorization. While factorization-based methods~\cite{ncf} disregard the temporal dynamics of interaction histories, GRU4Rec$^{+}$ utilize RNNs to incorporate the temporal dynamics. 
However, for datasets with longer sequences, the recurrent neural layers suffer from the vanishing gradient problem.
To alleviate this problem, HPMN introduced a hierarchical memory network to the recurrent layers to memorize the user's long-term interest.
On the other hand, BERT4Rec utilized the self-attention layer to globally attend all items within the sequence. 
Apart from the sequential recommendation literature, we consider MLP-based architectures (i.e., MLP-Mixer~\cite{tolstikhin2021mlpmixer} and gMLP~\cite{liu2021pay_attention}), which integrate the temporal dynamics with a simple MLP layer.

For POP and NCF we report the results from~\cite{bert4rec}. 
For GRU4Rec$^{+}$ and HPMN we use the official code while modifying the prediction module identical to our settings. 
For BERT4Rec, MLP-Mixer, and gMLP, the results are based on our reproduced implementations.

\noindent
\\
\textbf{Training} For the training procedure, we adopted the masked language modeling (MLM), a conventional training objective for bidirectional sequential models~\cite{devlin2018bert, bert4rec}. Specifically, we randomly masked tokens within the given sequence of historical behavior of a user and trained the model to predict the original item of the masked token based on its context. 

\noindent
\\
\textbf{Implementation Details} For fair comparison, we set GRU4Rec$^{+}$ and HPMN by following their optimal configurations where the number of layers are fixed as $L=6$.
For BERT4Rec, MLP-Mixer, gMLP, and \ourmodel{}, we use the optimal configurations of BERT4Rec. We trained all models by changing the learning rate from $\{1e-3, 5e-4, 3e-4, 1e-4\}$, weight decay from $\{1e-3, 1e-4, 1e-5\}$ and report the best results averaged ove three random seeds. 
Unless specified, we conducted experiments under the same experimental settings for all models and datasets, as follows: 

\begin{itemize}
    \setlength\itemsep{0.5em}
    \item \textit{Architecture}: We set the number of layers $L=2$ and input hidden dimension $d_h=256$. 
    We closely follow the implementation details in~\cite{tolstikhin2021mlpmixer} and set the hidden dimension $d_s$ of token-mixer as half the input hidden dimension $d_h$, i.e., $d_s = \frac{1}{2}d_h = 128$. 
    For the hidden dimension $d_c$ of channel-mixer, we scale the dimension for each model to minimize the difference of the total number of parameters between models.
    We use $d_c = 3d_h = 768$ for BERT4Rec, $d_c = 4d_h = 1024$ for gMLP, and $d_c = \frac{6}{k_c+1}d_h$ for \ourmodel{}.
    When $k_c=1$, the channel-mixing layer of \ourmodel{} is identical to the point-wise feed forward network of BERT4Rec. \\[0.5em]
    Following~\cite{bert4rec}, we initialized all parameters with a truncated normal distribution ranging from $[-0.02, 0.02]$ only except for those of gMLP's spatial gating unit. For the spatial gating unit, the weights are initialized nearly to zero and biases to one, in order to ensure the training stability as in~\cite{liu2021pay_attention}.
    
    \item \textit{Optimizer}: We used Adam~\cite{kingma2014adam} optimizer and trained for 200 epochs with cosine learning rate decay~\cite{loshchilov2016cosine}. We set the batch size as $256$ and dropout rate as $0.2$.

    \item \textit{Sequence Length}: Following~\cite{sasrec, bert4rec}, we set the maximum sequence length as $s=50$ for Beauty and Steam, and $s=200$ for ML-1m and ML-20m datasets. For XLong dataset, we set $s$ as 1000, following from~\cite{ren2019lifelong}. The mask proportion $\rho$ is fixed to $\rho = 0.6$ for Beauty, $\rho = 0.4$ for Steam, $\rho = 0.2$ for ML-1m, ML-20m, and $\rho = 0.1$ for XLong dataset.
\end{itemize}

\noindent
\\
\textbf{Evaluation} For evaluation, we adopted the next item recommendation (i.e., leave-one-out evaluation) task. For each user, we set the last item of the interaction sequence as the test data and the second-last item as the validation data. Here, the goal is to rank the ground-truth item higher than the other items. For fair comparison, we pair the ground-truth item with 100 negative items that are sampled according to their popularity following~\cite{ncf, sedhain2015autorec, sasrec, bert4rec}.

\noindent
\\
\textbf{Metrics}
When comparing among models, we are interested in two primary quantities: (1) the recommendation accuracy and (2) computational cost, which are major concerns of a recommendation system. To measure the recommendation accuracy, we commonly employ TOP-N metrics, \textit{Hit Ratio} (HR) and \textit{Normalized Discounted Cumulative Gain} (NDCG). Throughout this section, we compare HR@n and NDCG@n with rank $n$ set to $\{1,10\}$. To indicate the computational cost, we denote the number of parameters for the encoding layer (Params), the GPU memory requirements (VRAM) and the FLoating point Operations Per Second (FLOPs).

\subsection{Impact of Interaction Orders} \label{section:analysis_interaction_order}
Here, we empirically study the importance of the interaction order for the token- and channel-mixing layer. Throughout this section, we use ML-1m dataset for the evaluation.

\begin{table}[h]
\caption{Effects of the token interaction orders on ML-1m.}
\begin{tabular}{l l l c c}
\toprule
&\\[-3ex]               
Model               & $k_s$     & $k_c$     & Params (M)     & NDCG@10\\[0.2ex]
\hline\\[-2ex]
                    & 1         & 1         & 0.89           & \underline{0.4805}   \\
\ourmodel{}         & 2         & 1         & 0.89           & 0.4782    \\
                    & 3         & 1         & 0.89           & 0.4674    \\[0.2ex]
\hline
&\\[-2ex]
BERT4Rec            & -         & -         & 1.31           & \textbf{0.4964} \\[0.2ex]
\bottomrule
\end{tabular}
\label{table:token_comparison}
\end{table}

\begin{table*}[t]
\caption{Performance of different recommendation models on next-item prediction. The results are averaged over three random seeds. Bold scores indicate the best model for each metric and underlined scores indicate the second best model.}
\begin{tabularx}{\textwidth}{@{}YXYYYYYYYYYYYY@{}} 
\toprule
&\\[-3ex]
\multirow{2}{*}{Datasets} 
& \multirow{2}{*}{Metric}   & \multirow{2}{*}{POP}  & \multirow{2}{*}{NCF} & \multirow{2}{*}{GRU4Rec$^{+}$} & \multirow{2}{*}{HPMN} 
& \multirow{2}{*}{BERT4Rec} & \multirow{2}{*}{gMLP} & \multirow{2}{*}{MLP-Mixer}& \multicolumn{2}{c}{\ourmodel{}$_{(k_s, k_c)}$}  \\
\cmidrule(lr){10-11} 
&&&&&&&&& $(1, 2)$ & $(1, 3)$ \\[0.2ex]
\hline
&\\[-2ex]
\multirow{5}{*}{Beauty}
&Prms(M) & 0.0         & -           & 1.18     & 1.18    & 1.31      & 0.79     & 0.81     & 0.81               & 0.81   \\
&FLOPs(M)& 0.0         & -           & 14.58    & 7.60    & 16.38     & 10.90    & 11.95    & 11.95              & 11.95  \\
&HR@1    & 0.0077      & 0.0407      & 0.0651   & 0.0792  & 0.0988    & 0.0959   & 0.1068   & \textbf{0.1104}    & \underline{0.1102}  \\
&HR@10   & 0.0762      & 0.2142      & 0.2676   & 0.2803  & 0.3272    & 0.3217   & 0.3244   & \textbf{0.3294}    & \underline{0.3278}  \\
&NDCG@10 & 0.0349      & 0.1124      & 0.1479   & 0.1518  & 0.2000    & 0.1950   & 0.2029   & \textbf{0.2076}    & \underline{0.2073}  \\[0.2ex]
\hline
&\\[-2ex]
\multirow{5}{*}{Steam}
&Prms(M) & 0.0        & -            & 1.18     & 1.18        & 1.31            & 0.79     & 0.81     & 0.81                 & 0.81   \\
&FLOPs(M) & 0.0       & -            & 12.46    & 5.48        & 14.26           & 8.77     & 9.82     & 9.83                 & 9.83  \\
&HR@1     & 0.0159 	  & 0.0246       & 0.0832   &0.0986       & \textbf{0.1368} & 0.1138   & 0.1204   & 0.1278               & \underline{0.1305}    \\
&HR@10     & 0.1389      & 0.2169      & 0.3685   &0.4011       & \textbf{0.4560} & 0.4227   & 0.4245   & \underline{0.4430}& 0.4413              \\
&NDCG@10   & 0.0665      & 0.1026      & 0.2153   & 0.2315      & \textbf{0.2761} & 0.2476   & 0.2525   & 0.2635            & \underline{0.2657} \\[0.2ex]
\hline
&\\[-2ex]
\multirow{5}{*}{ML-1m}
&Prms(M) & 0.0         & -           & 1.18     & 1.18         & 1.31            & 0.87     & 0.89      & 0.89    & 0.89       \\
&FLOPs(M)& 0.0         & -           & 47.36    & 19.44        & 60.69           & 35.67    & 36.82     & 36.83   & 36.83     \\
&HR@1     & 0.0141     & 0.0397      & 0.2235   & 0.2631       & \textbf{0.3018} & 0.2823   & 0.2864    & \underline{0.2908}  & 0.2885     \\
&HR@10    & 0.1358      & 0.3477      & 0.6374   & 0.6612       & \textbf{0.7128} & 0.6958   & 0.6992    & \underline{0.7009}  & 0.7039     \\
&NDCG@10  & 0.0621      & 0.1640      & 0.4151   & 0.4468       & \textbf{0.4964} & 0.4781   & 0.4805    & \underline{0.4873}  & 0.4853     \\[0.2ex]
\bottomrule
\end{tabularx}
\label{table:small_scale_result}
\end{table*}

\noindent
\textbf{Order of token interaction} Table~\ref{table:token_comparison} shows the comparison of \ourmodel{} to other methods with respect to various token interaction orders ranging from $k_s \in \{1,2,3\}$. 
Comparing the results among MOI-Mixer variants, MOI-Mixer$_{(1,1)}$ (i.e., MLP-Mixer) shows the most competitive performance compared to BERT4Rec, with a $1.6\%$ drop in NDCG@10. We observe that a simple first-order token interaction performs the best, while more complex token-wise interactions degrade the recommendation accuracy of \ourmodel{}.

\begin{table}[h]
\caption{Effects of $d_s$ for the token-mixer in \ourmodel{}$_{(1,1)}$. The results indicate NDCG@10 for ML-1m.}
    \begin{tabular}{l c c c c c c}
\toprule
&\\[-3ex]
Model           & $16$   & $32$   & $64$   & $128$  & $256$  & $512$ \\[0.5ex]
\hline &\\[-2ex]
MOI$_{(1,1)}$  & 0.4485 & 0.4652 & 0.4737 & \underline{0.4805} & 0.4791 & \textbf{0.4814} \\
\bottomrule
\end{tabular}
\label{table:token_hidden_dim}
\end{table}

We also observe that \ourmodel{}$_{(1,1)}$ works reasonably well with a wide range of token hidden dimensions as described in Table \ref{table:token_hidden_dim}.
Although ML-1m's maximum sequence length is set to $s=200$, utilizing a bottleneck architecture of $d_s=128$ showed subtle performance difference to $d_s=512$. Moreover, in the extreme case of encoding the interactions with $d_s=16$, the performance dropped only by $3.2\%$. 

From these experimental results, we claim that the token-mixing operation of sequential recommendation may not require complex interactions and a simple interaction may be sufficient.

\begin{table}[h]
\caption{Effects of the channel interaction orders on ML-1m.}
\begin{tabular}{l l l c c}
\toprule
&\\[-3ex]               
Model               & $k_s$     & $k_c$     & Params (M)     & NDCG@10\\[0.2ex]
\hline\\[-2ex]
                    & 1         & 1         & 0.89              & 0.4805    \\
\ourmodel{}         & 1         & 2         & 0.89              & \underline{0.4873} \\
                    & 1         & 3         & 0.89              & 0.4853    \\[0.2ex]
\hline
&\\[-2ex]
BERT4Rec            & -         & -         & 1.31           & \textbf{0.4964} \\[0.2ex]
\bottomrule
\end{tabular}
\label{table:channel_comparison}
\end{table}

\noindent
\textbf{Order of channel interaction}
Table~\ref{table:channel_comparison} reports the results of \ourmodel{} for different channel interaction orders $k_c \in \{1,2,3\}$.
We discovered that increasing the order of channel interaction from $k_c=1$ to $k_c=2$ boosts the performance by $0.7\%$. 
A possible explanation is that \ourmodel{} benefits from the attained fine-grained channel representations by explicitly expressing the multiplicative channel interaction as in~\cite{FiBiNet}. 
Interestingly, when we raise the order of channel interaction from $k_c=2$ to $k_c=3$, it does not improve the performance of our model while the results are similarly good, with a performance drop of $0.2\%$. 

In summary, we found that the high-order interaction was helpful in the channel-mixing layer but was not a necessary component in the token-mixing layer. 
Therefore, unless specified, we will use \ourmodel{}$_{(1,2)}$, which showed superior performance over \ourmodel{}$_{(1,3)}$, as our default architecture throughout the paper.
We provide additional analysis regarding the impact of different combinations of token- and channel-mixing interaction order on the performance of \ourmodel{} in the Appendix A.1.

\subsection{Results} \label{section:small_scale}

Here, we empirically study the performance of our method on various types of experimental settings. 
We categorize the datasets as small-scale, large-scale, and long-sequence, to evaluate our model. For all evaluations, we omit NDCG@1 as it is identical to HR@1.

\noindent
\\
\textbf{Small-scale} Table~\ref{table:small_scale_result} summarizes the results of all models on Beauty, Steam and ML-1m. First, by comparing POP with NCF, we observe that reflecting the user's personal preferences is beneficial. Second, explicitly modeling the temporal dynamic with recurrent layers (i.e., GRU4Rec$^+$, HPMN) outperform the matrix factorization-based method (i.e., NCF). Third, BERT4Rec significantly outperforms the models with the recurrent layers verifying the effectiveness of the self-attention layers in capturing the user's preference.

Among the MLP-based methods, \ourmodel{} shows the best performance for all datasets. For Beauty, \ourmodel{} attained state-of-the-art performance with $0.76$\% higher NDCG@10 results, $38.2$\% fewer parameters and $27$\% fewer FLOPs compared to the Transformer-based model (i.e., BERT4Rec). 
Though \ourmodel{} was not the best performing architecture for Steam and ML-1m, it achieves competitive results compared to the Transformer-based model (i.e., BERT4Rec). More specifically, compared to MLP-Mixer, \ourmodel{} shows $1.1$\% and $0.68$\% improvements in NDCG@10 for Steam and ML-1m, respectively, with identical computational cost. 
It is interesting to note that MOI-Mixer$_{(1,2)}$ achieved superior performance to MOI-Mixer$_{(1,3)}$ for Beauty and ML-1m but not for Steam. We conjecture that the Steam dataset requires more complex features than other datasets when expressing the tokens.

In summary, we demonstrate the importance of including high-order channel interactions to obtain the effective representation for the sequential recommendation datasets at a relatively small scale.

\begin{table}[h]
\caption{The number of parameters of each model with a different number of layers and hidden dimensions.}
\begin{tabular}{lcccccc}
\toprule
&\\[-3ex]
\multirow{3}{*}{Name}   & \multicolumn{2}{c}{Dimensions} & \multicolumn{4}{c}{Params (M)}  \\
                          \cmidrule(lr){2-3}                \cmidrule(lr){4-7} 
                        & \multirow{2}{*}{$L$} & \multirow{2}{*}{$d_h$} & \multirow{2}{*}{BERT4Rec}   & \multirow{2}{*}{gMLP}  & MLP   & MOI \\
                        &     &                           &             &                                                      & Mixer & Mixer \\
\hline
&\\[-2ex]
\multirow{2}{*}{Base}   & 12  & 512                      & -        & 19.4  & 20.1  & 20.1        \\
                        & 8   & 512                      & 21.0     & 12.9  & 13.4  & 13.4        \\
\hline
&\\[-2ex]
\multirow{2}{*}{Small}  & 4   & 512                      & 10.5     & 6.5   & 6.7   & 6.7        \\
                        & 4   & 256                      & 2.6      & 1.8   & 1.8   & 1.8        \\
\hline
&\\[-2ex]
\multirow{1}{*}{Micro}  & 2   & 256                      & 1.3      & 0.9   & 0.9   & 0.9        \\
\bottomrule
\end{tabular}
\label{table:large_scale_result}
\end{table}

\noindent
\textbf{Large-scale} Here, we study the scalability of \ourmodel{} compared to BERT4Rec, gMLP, and MLP-Mixer in a larger dataset, i.e., ML-20m.
Table~\ref{table:large_scale_result} reports the number of parameters for each model by changing the number of layers and hidden dimensions where the configurations are adopted from~\cite{devlin2018bert}.

\begin{table}[h]
\caption{Performance comparison on ML-20m dataset using the largest model configuration.}
\begin{tabular}{l c c c c}
\toprule
&\\[-3ex]
Model               & Prms (M)   & FLOPs (M)   & HR@10     & NDCG@10\\[0.2ex]
\hline
&\\[-2ex]
BERT4Rec            & 21.0        & 906.4        & \textbf{0.7756}   & \textbf{0.5647}  \\
gMLP                & 19.4        & 806.3        & 0.7712   & \underline{0.5622}      \\
MLP-Mixer           & 20.1        & 882.1        & 0.7684   & 0.5580  \\
MOI-Mixer& 20.1        & 882.9        & \underline{0.7730}   & 0.5620  \\
\bottomrule
\end{tabular}
\label{table:large_scale_result}
\end{table}

Table~\ref{table:large_scale_result} summarizes the performance of the models with their largest configurations.
We found that the performance difference was marginal, showing the largest difference of NDCG@10 was only $0.67$\%.
Among the MLP-based architectures, \ourmodel{} shows competitive scalability against gMLP, which is the state-of-the-art MLP-based architecture in natural language processing tasks~\cite{liu2021pay_attention}.

Fig.~\ref{large_scale_plot} summarizes the performance vs parameter curve for each model. 
Comparing BERT4Rec with the MLP-based architectures, the performance gap reduces as the model scale increases. 
Moreover, \ourmodel{} and gMLP were able to surpass MLP-Mixer at all scales.
This verifies that an explicit high-order term is consistently beneficial for learning sequential patterns.

In summary, similar to the findings from the small-scale datasets, the explicit high-order feature interactions were also beneficial for the large-scale dataset. 

\vspace{3mm}
\pgfplotsset{compat=1.17}
\begin{tikzpicture}
\begin{groupplot}[
      group style={group size=1 by 1, horizontal sep=1.2cm},
      width=8cm, height=5cm,]
    \nextgroupplot[
        every x tick label/.append style={font=\scriptsize\color{gray!80!black}},
        xmin=0, xmax=22,
        xlabel={\small Params (M)},   
        every y tick label/.append style={font=\scriptsize\color{gray!80!black}},
        ymin=0.505, ymax=0.572,
        ylabel={\small NDCG@10},
        ylabel style={yshift=-1mm},
        xtick={1, 7, 13, 21},
        grid=both,
        tick align=outside,
        tick pos=left,
        ytick={0.51, 0.52, 0.53, 0.54, 0.55, 0.56, 0.57},
        legend style={at={($(0,0)+(1cm,1cm)$)},legend columns=4,fill=none,draw=black,anchor=center,align=center, font=\small},
        legend to name = models]
    \addplot[color=customblue,
             mark=*,
             mark size=1.3] 
    coordinates{(1.3,0.5320) (2.6,0.5421) (10.5,0.5564) (21.0,0.5647)};
    \addplot[color=customyellow,
             mark=triangle*,
             mark size=1.5] 
    coordinates{(0.9,0.5107) (1.8,0.5313) (6.5,0.5436) (12.9,0.5566) (19.4,0.5620)};
    \addplot[color=customgreen,
             mark=square*,
             mark size=1.3] 
    coordinates{(0.9,0.5180) (1.8,0.5244) (6.7,0.5366) (13.4,0.5502) (20.1,0.5580)};
    \addplot[color=customred,
             mark=square*,
             mark size=1.3] 
    coordinates{(0.9,0.5224) (1.8,0.5324) (6.7,0.5420) (13.4,0.5548) (20.1,0.5622)};
    \addlegendentry{BERT4Rec};    
    \addlegendentry{gMLP};
    \addlegendentry{MLP-Mixer};
    \addlegendentry{MOI-Mixer};
    \coordinate (c1) at (rel axis cs:0,1);
    \coordinate (c2) at (rel axis cs:1,1);
    \end{groupplot}
    \coordinate (c3) at ($(c1)!.5!(c2)$);
    \node[above] at (c3 |- current bounding box.north)
    {\pgfplotslegendfromname{models}};
\end{tikzpicture}
\vspace*{-5mm}
\captionsetup{justification=centering}
\captionof{figure}{Performance-parameter trade-offs at a different scale, evaluated on ML-20m dataset.}
\label{large_scale_plot}


\begin{table}[h]
\caption{Performance of recommendation models on XLong dataset. The maximum sequence length $s$ is set to 1000.}
\begin{tabular}{l c c c c}
\toprule
&\\[-3ex]
Model               & Prms (M)  & VRAM (GB)    & FLOPs (M)      & NDCG@10\\[0.2ex]
\hline
&\\[-2ex]
GRU4Rec$^{+}$       & 0.14        & 4.37         & 14.75        & 0.1249     \\
HPMN                & 0.14        & 3.54         & 6.07         & 0.2657      \\
BERT4Rec            & 0.08        & 42.09        & 67.28        & 0.4852  \\
gMLP                & 2.05        & 13.14        & 35.40        & 0.4902      \\
MLP-Mixer           & 0.30        & 13.32        & 13.04        & \underline{0.4914}  \\
MOI-Mixer           & 0.30        & 13.52        & 13.08        & \textbf{0.4938}  \\
\bottomrule
\end{tabular}
\label{table:long_sequence_result}
\end{table}

\noindent
\textbf{Long Sequence} Table~\ref{table:long_sequence_result} summarizes the results of GRU4Rec$^{+}$, HPMN, BERT4Rec, gMLP, MLP-Mixer and \ourmodel{} for the XLong dataset. Our goal is to investigate the performance and capacity of each model for modeling lifelong sequences. To fully compare the computational complexity of each model, we set $d_h=64$, the batch size as 128, and measure all metrics on two RTX 3090 24GB GPUs. Our observations from the results are as follows:

\hspace{0.2cm} \ourmodel{} achieves the best performance while maintaining identical computation cost of MLP-Mixer, confirming the benefit of high-order interaction for long sequences.
\ourmodel{} only required $85.4\%$ fewer parameters and $63.1\%$ fewer FLOPS compared to gMLP, which shows the lowest performance out of all MLP-based models.
Moreover, \ourmodel{} required $67.9\%$ fewer VRAM and $80.6\%$ fewer FLOPS than BERT4Rec with $0.86\%$ higher NCDG@10 results.

\begin{figure}[h]
    \begin{center}
    \includegraphics[trim={2.3cm 17cm 10.7cm 3cm}, width=1.0\linewidth]{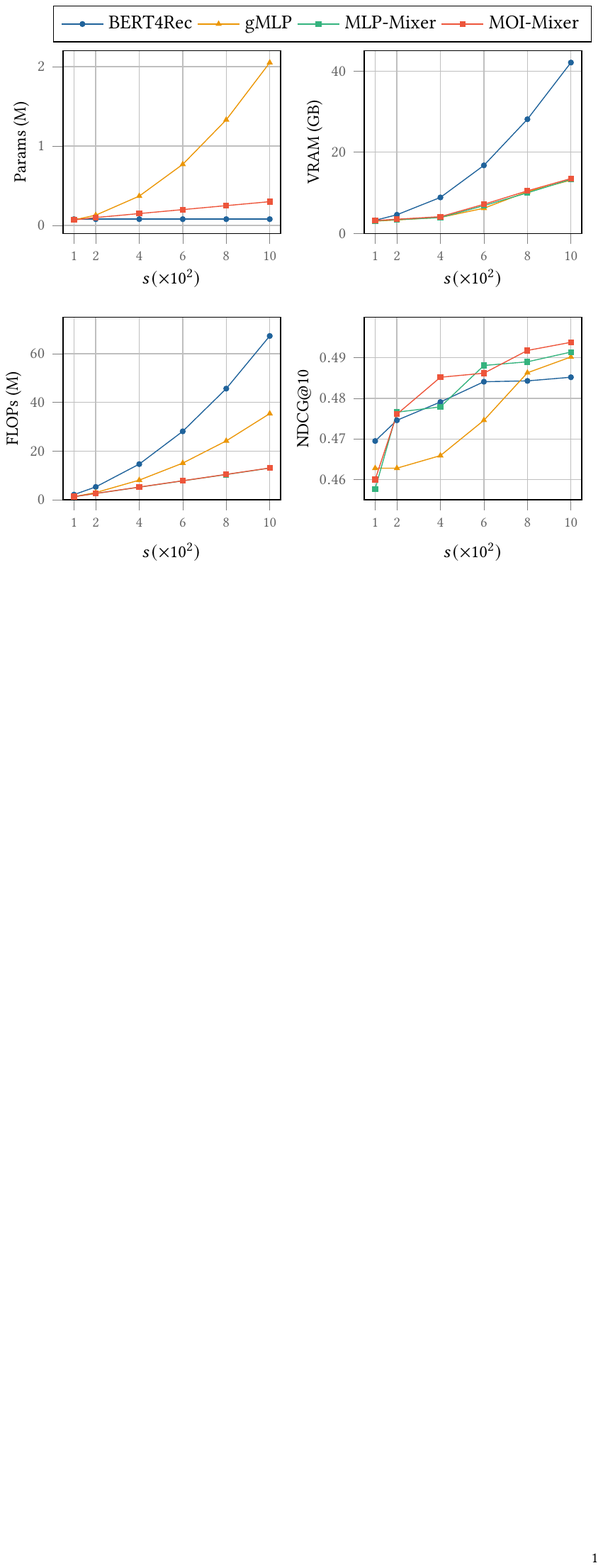}
    \end{center}
    \captionsetup{justification=centering}
    \caption{Comparison of different recommendation models in terms of the computational cost and performance. The results are obtained from XLong dataset.}
    \label{fig:long_sequence}
\end{figure}

\hspace{0.2cm} Fig.~\ref{fig:long_sequence} illustrates the computation cost and NDCG@10 of the compared models by changing the sequence length $s$ from $100$ to $1000$.
We find that all models yield better recommendation accuracy when given longer sequences, demonstrating the importance of integrating long sequence of interactions.
However, in BERT4Rec, the required VRAM and FLOPs grow exponentially with longer sequences due to the attention weight, which is quadratic to a sequence length, and in gMLP, the total number of parameters and FLOPs grow exponentially.
Such properties limit the applicability of BERT4Rec and gMLP for real-world tasks requiring modeling long sequences.
In contrast, MLP-Mixer and \ourmodel{} efficiently processes long sequences in linear computational complexity. 

\subsection{Ablation studies} \label{section:ablation}

Table~\ref{table:ablation} reports the ablation study of our base model and a summary of our preliminary exploration.

\begin{table}[h]
\caption{Ablation study on the architectural components of \ourmodel{}, evaluated on ML-1m dataset.}
\begin{tabular}{l l c}
\toprule
&\\[-3ex]
Ablation & Variant              & NDCG@10 \\[0.2ex]
\hline 
&\\[-2ex]
Baseline & MOI-Mixer$_{(1,2)}$  & 0.4873 \\
\hline 
&\\[-2ex]
\multirow{3}{*}{Norm-Type}      & LayerNorm $\rightarrow$ None         & 0.4848 \\
                                & LayerNorm $\rightarrow$ L2-Norm      & 0.4767 \\
                                & LayerNorm $\rightarrow$ Layer-Scale  & 0.4862 \\
\hline 
&\\[-2ex]
\multirow{2}{*}{Norm-Location}  & Before-Activation  & 0.4840 \\
                                & After-Activation   & 0.4859 \\
\hline 
&\\[-2ex]
Token-mixer                     & MLP $\rightarrow$ Linear  & 0.4843 \\
\hline 
&\\[-2ex]
\multirow{1}{*}{Embedding}      & (+) Position Embedding       & 0.4876 \\
\bottomrule
\end{tabular}
\label{table:ablation}
\end{table}

\noindent
\\
\textbf{Normalization} To see how different types of normalization can affects our model performance, we experimented using (i) no normalization, (ii) L2 normalization as in~\cite{pmlr2020lowrankbilinear} and (iii) Layer Scale~\cite{touvron2021going}. It was interesting to observe that applying L2 normalization degrades the performance more than when applying no normalization at all. Also, while Layer Scale was shown to improve Transformer architectures, \ourmodel{} does not benefit much from it. We also performed ablation studies to see whether the location of the normalization layer is critical. Through preliminary experiments, we observed that applying the normalization layer right after computing the Hadamard Product was the most effective. 


\noindent
\\
\textbf{Token-mixer} Replacing the MLP layer with a a single linear layer for the token-mixing component also gives a good performance, with only a $0.3$\% decrease in NDCG@10. However, such replacement is not desirable since it creates a quadratic term for the sequence length, restricting the scalability of \ourmodel{}. 

\noindent
\\
\textbf{Embedding Layer} Following the implementation details of Transformer, we also evaluated the performance of \ourmodel{} with additionally applying positional embeddings. The position embedding did not bring gain tio \ourmodel{} since the MLP layer is already sensitive to the order of the sequences.

\section{Conclusion and Discussion} \label{section:conclusion}

This paper proposed MOI-Mixer, which aims to leverage high-order interactions for MLP-based models in sequential recommendation systems.
We claim that Transformer and existing MLP-based models differ in performance due to the absence of an explicit high-order term. 
Thus, we introduce a novel MOI layer which is capable of modeling arbitrary multi-order interactions among the given input features. 
Experimental results on five real-world datasets show that integrating a high-order term in MLP-based models is consistently beneficial. 
We also show that \ourmodel{} is computationally efficient in processing long-sequence behavior data compared to the state-of-the-art model~\cite{bert4rec}. 

\hspace{0.2cm} Interesting future work includes applying \ourmodel{} to various other fields such as computer vision and natural language processing, where the simple MLP-based models have shown competitive results. We hope that our work opens up the potentials for research on MLP-based models in sequential recommendation tasks. 

\appendix
\section{APPENDIX: ADDITIONAL EXPERIMENTS}\label{appendix}

\subsection{Analysis on the interaction orders}
The following table reports the results of \ourmodel{} by varying the order of the token- and the channel-mixing layers.

\begin{table}[h]
\caption{\ourmodel{}'s performance on ML-1m dataset by varying the combinations of the interaction orders.}

\begin{tabular}{l c c c c}
\toprule
&\\[-3ex]
$k_s$  \textbackslash $\text{ }k_c$ & 1   & 2   & 3     & 4\\[0.2ex]
\hline
&\\[-2ex]
1           & 0.4805      & \textbf{0.4873}  & 0.4853               & \underline{0.4855}  \\
2           & 0.4782      & 0.4846           & 0.4835               & 0.4812  \\
3           & 0.4674      & 0.4743           & 0.4729               & 0.4711  \\
4           & 0.4653      & 0.4738           & 0.4708               & 0.4695  \\
\bottomrule
\end{tabular}
\label{table:order_result}
\end{table}

In this table, we observe that (i) increasing interaction order between the tokens consistently degrades the performance regardless of the channel's interaction order and (ii) the second-order channel interaction performed the best where further increasing the order of channel interactions weakens the performance in \ourmodel{}.

\clearpage


\bibliographystyle{ACM-Reference-Format}
\bibliography{reference}
\clearpage

\end{document}